\documentclass[runningheads]{llncs}
\usepackage{amssymb}
\usepackage{amsmath}
\usepackage{graphicx}
\usepackage{tikz}
\usepackage[export]{adjustbox}

\usepackage{graphicx}
\usepackage{tabularx}
\usepackage{booktabs}
\usepackage{dcolumn}
\usepackage{hyperref}
\usepackage{subcaption}

\usepackage{makecell}

\DeclareMathOperator*{\argmax}{\arg\max}

\begin{document}
\title{Recurrent Vision Transformer for Solving Visual Reasoning Problems}
%
%
\author{Nicola Messina \and Giuseppe Amato \and Fabio Carrara \and Claudio Gennaro \and Fabrizio Falchi}
\authorrunning{N. Messina et al.}
%
\institute{Institute of Information Science and Technologies (ISTI), Italian National Research Council (CNR), Via G. Moruzzi 1, 56124 Pisa, Italy\\
\email{name.surname@isti.cnr.it}}
%
\maketitle              
\begin{abstract}
Although convolutional neural networks (CNNs) showed remarkable results in many vision tasks, they are still strained by simple yet challenging visual reasoning problems. Inspired by the recent success of the Transformer network in computer vision, in this paper, we introduce the Recurrent Vision Transformer (RViT) model. Thanks to the impact of recurrent connections and spatial attention in reasoning tasks, this network achieves competitive results on the \textit{same-different} visual reasoning problems from the SVRT dataset. The weight-sharing both in spatial and depth dimensions regularizes the model, allowing it to learn using far fewer free parameters, using only 28k training samples. A comprehensive ablation study confirms the importance of a hybrid CNN + Transformer architecture and the role of the feedback connections, which iteratively refine the internal representation until a stable prediction is obtained. In the end, this study can lay the basis for a deeper understanding of the role of attention and recurrent connections for solving visual abstract reasoning tasks. The code for reproducing our results is publicly available here: [hidden for double-blind review].


\end{abstract}
\section{Introduction}

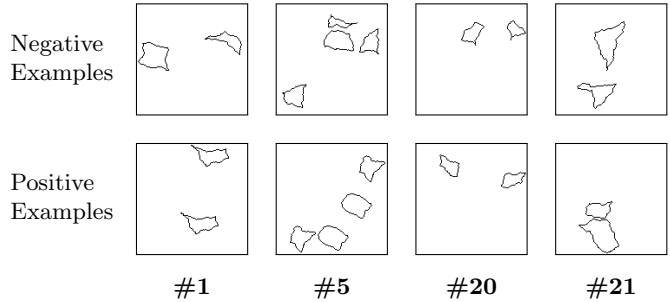
\begin{figure}[t]
\centering
\begin{tikzpicture}[picture format/.style={inner sep=5pt,}]

  \node[picture format]                   (A1)               {\includegraphics[width=0.12\textwidth,frame]{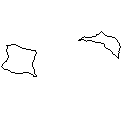}};
  \node[picture format,anchor=north]      (B1) at (A1.south) {\includegraphics[width=0.12\textwidth,frame]{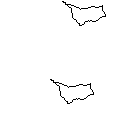}};

  \node[picture format,anchor=north west] (A2) at (A1.north east) {\includegraphics[width=0.12\textwidth,frame]{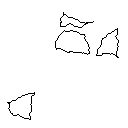}};
  \node[picture format,anchor=north]      (B2) at (A2.south)      {\includegraphics[width=0.12\textwidth,frame]{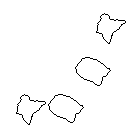}};

  \node[picture format,anchor=north west] (A3) at (A2.north east) {\includegraphics[width=0.12\textwidth,frame]{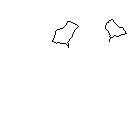}};
  \node[picture format,anchor=north]      (B3) at (A3.south)      {\includegraphics[width=0.12\textwidth,frame]{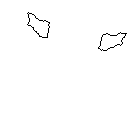}};
  
  \node[picture format,anchor=north west] (A4) at (A3.north east) {\includegraphics[width=0.12\textwidth,frame]{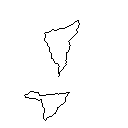}};
  \node[picture format,anchor=north]      (B4) at (A4.south)      {\includegraphics[width=0.12\textwidth,frame]{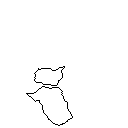}};


  \node[anchor=north] (C1) at (B1.south) {\bfseries \#1};
  \node[anchor=north] (C2) at (B2.south) {\bfseries \#5};
  \node[anchor=north] (C3) at (B3.south) {\bfseries \#20};
  \node[anchor=north] (C4) at (B4.south) {\bfseries \#21};
  
  \node[anchor=east,align=left] (C1) at (A1.west) {Negative\\Examples};
  \node[anchor=east,align=left] (C2) at (B1.west) {Positive\\Examples};

\end{tikzpicture}
\caption{Positive and negative examples from the four considered SVRT problems: P.1: same shapes; P.5: two twisted pairs of same shapes; P.20: same shapes reflected along an unknown symmetry axis; P.21: same shapes but rotated and scaled.}
\label{fig:svrt-examples}
\end{figure}

Deep learning methods largely reshaped classical computer vision, solving many tasks impossible to face without learning representations from data.
The well-established convolutional neural network (CNN) architecture for image processing obtained state-of-the-art results in many computer vision tasks, such as image classification \cite{foret2020sharpness,xie2017aggregated}, or object detection \cite{ren2015faster,redmon2018yolov3,ciampi2020virtual}. Recently, a novel promising architecture took hold in the field of image processing: the Transformer. Initially developed for solving natural language processing tasks, it found its way into the computer vision world, capturing the interest of the whole community. 
These Transformer-based architectures already proved their effectiveness in many image and video processing tasks \cite{messina2020fine,messina2021transformer,dosovitskiy2020image,coccomini2021combining,bertasius2021space}. 
The Transformer's success is mainly due to the power of the self-attention mechanism, which can relate every visual token with all the others, creating a powerful relational understanding pipeline. 
In this paper, we aim at studying the relational understanding capabilities of Vision Transformers in the context of an apparently simple yet non-trivial task, called \textit{same-different} task. In short, the same-different task consists in understanding if two shapes in an image satisfy a certain rule. In the simpler case, the rule is merely that \textit{the two shapes must be equal}; however, the rule is not known a priori and must be internally understood from the provided positive and negative examples. An example is given in Figure \ref{fig:svrt-examples}.

This task can be framed as a binary classification problem, and it has been partially solved with state-of-the-art convolutional architectures, particularly with ResNets \cite{funke2021five,puebla2021can,borowski2019notorious,messina2021solving}. From these studies, it has been observed that (a) deep CNNs are needed, with lots of free parameters, to relate distant zones of the image in search of matching patterns, and (b) usually, a lot of data is needed to learn the underlying rule, while humans can spot it with only a few samples. Furthermore, some works \cite{Kar2019} emphasized the role of recurrent connections, which can iteratively refine the visual input until an optimal and stable conclusion is drawn. 
In the light of these observations, in this paper, we introduce a novel architecture, called Recurrent Vision Transformer (RViT), for solving the same-different problems. It is inspired by both the recent Vision Transformer (ViT) model \cite{dosovitskiy2020image} and by a recurrent version of the Transformer architecture called Universal Transformer \cite{dehghani2018universal}. The introduced architecture can understand and relate distant parts in the image using the powerful Transformer's attentive mechanism and iteratively refine the final prediction using feedback connections. 
Notably, we find that the base ViT model cannot learn any of the same-different tasks, suggesting that both a hybrid architecture (upstream CNN + downstream Transformer) and feedback connections can be the keys for solving the task. 


To summarize, the contribution of the paper is many-fold: (a) we introduce a novel architecture, called Recurrent Vision Transformer (RViT), a hybrid CNN + Transformer architecture for solving the challenging same-different tasks; (b) we compare the network complexity and accuracy with respect to other architectures on the same task, obtaining remarkable results with less free parameters and thus better data efficiency; (c) we qualitatively inspect the learned attention maps to understand how the architecture is behaving, and we provide a comprehensive study on the role of the recurrent connections.


\section{Related Work}

\paragraph{Vision Transformers}
The massive engagement of the Transformer architecture \cite{vaswani2017transformer} in the Natural Language Processing community grew at the point that it trespassed the boundaries of language processing, finding wide applications in computer vision.
In fact, it is possible to subdivide images into \textit{patches} which can be fed as input to a Transformer encoder for further processing; 
the self-attentive mechanism can discover long-range dependencies between image patches, overcoming the limits of the local processing performed by CNNs. 

On these simple concepts, several different architectures have been proposed. Some of these, like Cross Transformers \cite{doersch2020crosstransformers} or DETR \cite{carion2020end}, use the regular grid of features from the last feature map of a CNN as visual tokens. 
More recently, fully-transformer architectures, first among which ViT \cite{dosovitskiy2020image}, have taken root. For the first time, no convolutions are used to process the input image. In particular, the ViT architecture divides the image in patches using the grid approach; the RGB pixel values from every patch are concatenated, and they are linearly projected to a lower-dimensional space to be used as visual tokens. The BERT-like [CLS] token \cite{devlin2019bert} is then used as the classification head.
On the same wave of ViT, the TimeSformer \cite{bertasius2021space} redefined attention both in space and time to understand long-range space-time dependencies in videos.

\paragraph{Same-different Task}
Many tasks have been proposed in computer vision to tackle abstract visual reasoning abilities of machine learning models, like CLEVR and Sort-of-CLEVR \cite{Johnson2017CLEVR}, Raven's Progressive Matrices (RPM), or Procedurally Generated Matrices (PGMs) \cite{barrett2018abstractreasoning}.
In \cite{Fleuret2011}, the authors introduced the \textit{Synthetic Visual Reasoning Test} (SVRT) dataset, composed of simple images containing closed shapes. It was developed to test the relational and comparison abilities of artificial vision systems.
The work in \cite{Stabinger2016} first showed, using the SVRT dataset, that the tasks involving comparisons between shapes were difficult to solve for convolutional architectures like LeNet and GoogLeNet \cite{Szegedy2015GoogLeNet}. The authors in \cite{kim2018not} drawn a similar conclusion, introducing a variation of the SVRT dataset -- the Parametric SVRT (PSVRT) for solving some shortcomings of the SVRT dataset -- and concluding that the Relation Network \cite{Santoro2017RelationNetworks} is also strained on the same-different judgments. Similarly, \cite{puebla2021can} developed a more controlled visual dataset to evaluate the reasoning abilities of deep neural networks on shapes having different distributions.
The authors in \cite{borowski2019notorious,funke2021five} found that deep CNNs, like ResNet-50, can solve the SVRT problems even with a relatively small amount of samples (28k images). Similarly, the authors in \cite{messina2019testing,messina2021solving} demonstrated that also many other state-of-the-art deep learning architectures for classifying images (ResNet, DenseNets, CorNet) models can learn this task, generalizing to some extent. 
Recently, \cite{vaishnav2021understanding} discussed the important role of attention in deep learning models for addressing the same-different problems.

\paragraph{Recurrent Models}
Recurrent models -- LSTMs \cite{hochreiter1997long} and GRUs \cite{cho2014learning}, to name a few -- have been widely used for dealing with variable-length sequences, especially in the field of natural language processing. However, recently, many neuroscience and deep-learning works claimed the importance of recurrent connections outside the straightforward text processing, as they could have an essential role in recognition and abstract reasoning. The work in \cite{Kar2019} claimed that the visual cortex could be comprised of recurrent connections, and the visual information is refined in successive steps. Differently, many works in deep learning tried to achieve Turing-completeness by creating recurrent architectures with dynamic halting mechanisms \cite{graves2014neural,dehghani2018universal,banino2021pondernet}. Although our work does not include dynamic halting mechanisms, it partially embraces these ideas, experimenting with recurrent connections for iteratively refining the final prediction.

\section{The Recurrent Vision Transformer Model}
\begin{figure}[t]
  \centering
  \includegraphics[page=1,width=1\linewidth]{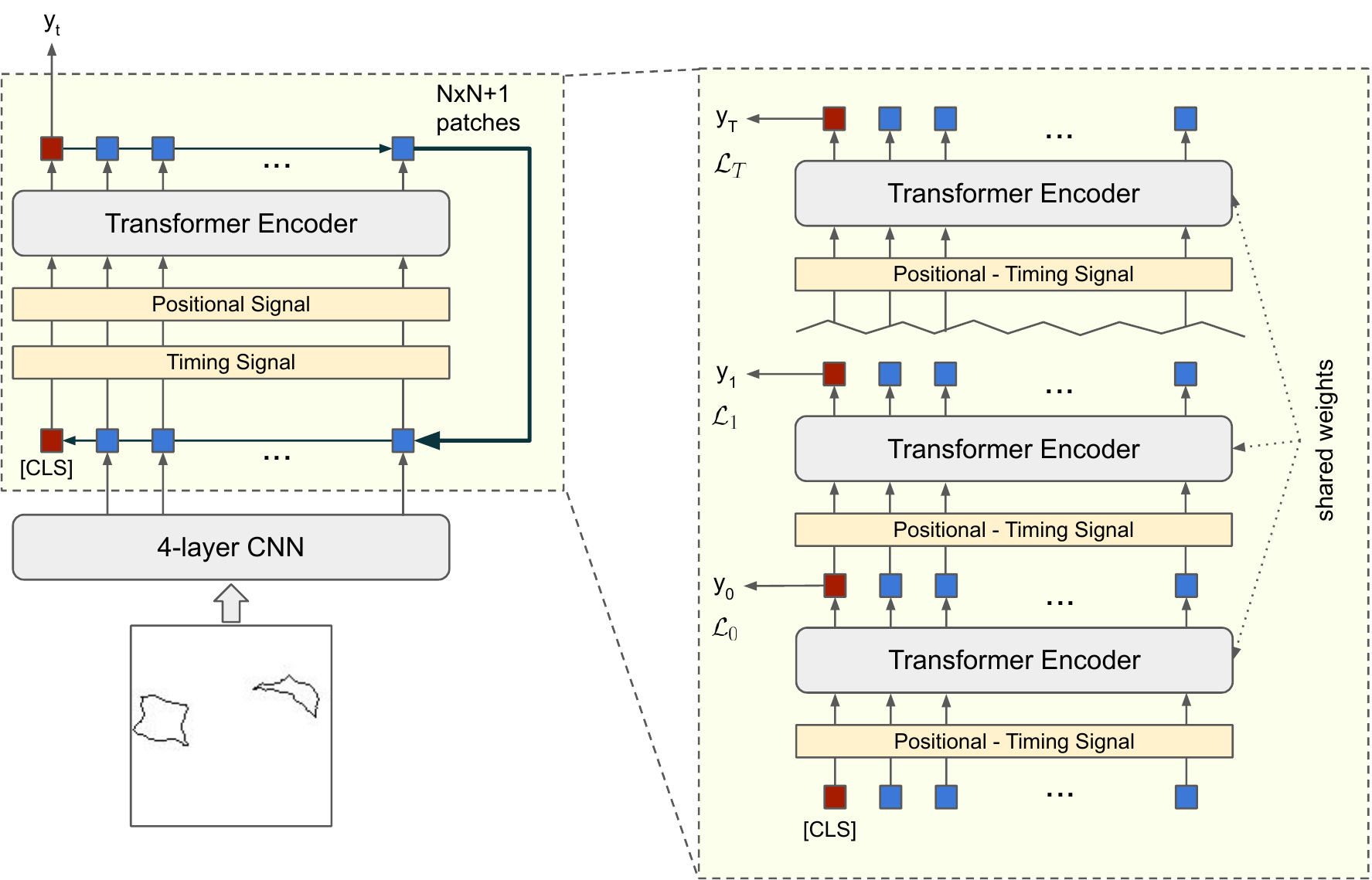}
  \caption{The RViT architecture. The image is processed by a 4-layer CNN, outputting a $8\times8$ grid of visual features. The CLS token is added to this set, and the tokens are processed multiple times by the recurrent transformer encoder module. At each time step, the binary cross-entropy loss is computed against the ground-truth labels.}
  \label{fig:architecture} 
\end{figure}

The proposed model is based on the recent Vision Transformer -- in particular, the ViT model \cite{dosovitskiy2020image}. The drawback of CNNs in solving the same-different problems is that sufficiently deep networks are needed to correlate distant zones in the image. The Transformer-like attention mechanism in ViT helps in creating short paths between image patches through the self-attention mechanism. Furthermore, inspired by the role of recurrent connections in the human's visual cortex \cite{Kar2019}, we modify the ViT Transformer encoder module by sharing the encoder weights among all the $T$ layers (i.e., along the depth dimension), effectively creating a recurrent Transformer encoder model, similar to \cite{dehghani2018universal}. 
This has the effect of sharing weights not only in the sequence dimension as in standard Transformers, but also in the depth dimension, further constraining the model complexity. As a feature extractor, we use a small upstream CNN that outputs $N \times N$ $D-$dimensional features used as visual tokens in input to the Transformer encoder.
The overall architecture is shown in Figure \ref{fig:architecture}.

By leveraging the recurrent nature of the architecture, we avoid explicitly tuning the depth of the network (i.e., the total number of recurrent iterations) by forcing the architecture to perform a prediction at each time step, using the CLS token. The most likely outcome among the predictions from all the time steps is then taken as the final prediction.
More in detail, the model comprises $T$ binary classification heads, one for each time step. During training, the binary cross-entropy loss at each time step is computed as
\begin{equation}
    \mathcal{L}_t = \text{BCE}(y_t, \hat{y}),
\end{equation}
where $y_t$ is the network output from the $t$-th time step, and $\hat{y}$ is the ground-truth value.
The various losses are then aggregated using the automatic loss-weighting scheme proposed in \cite{kendall2018multi}:
\begin{equation}
    \mathcal{L}_{\text{total}} = \sum_t^T \frac{1}{2} \left ( \frac{1}{e^{s_t}} \mathcal{L}_t + s_t \right ),
\end{equation}
where $s_t$ is a free scalar parameter that encodes the predicted uncertainty of the classification at the $t$-th time step, and the model automatically learns it during the training phase. We refer readers to \cite{kendall2018multi} for more detailed derivation and discussion.

During inference, the maximum-likelihood prediction is taken as the final network output. In particular, the time step $\bar{t}$ at which the network reaches the maximum confidence is the one where the output probability is farthest from the pure chance in a binary classification setup (p=0.5): 
\begin{equation}
\bar{t} = \argmax_t |y_t - 0.5|.
\end{equation}
At this point, the final output is simply $y = y_{\bar{t}}$.

\section{Experiments}
In this section, we briefly introduce the SVRT dataset used in the experiments, and we present and discuss the performance of the Recurrent Vision Transformer on these problems.

\subsection{Dataset}

In this work, we use the \textit{Synthetic Visual Reasoning Test} (SVRT) benchmark to test our proposed architecture. SVRT comprises 23 different sub-problems; each sub-problem comprises a set of positive and negative samples generated using a problem-specific rule. The objective of any classifier trained on a problem is to distinguish the positive and negative samples, and the only way to succeed is to discover the underlying rule. 

From previous works \cite{kim2018not,borowski2019notorious} it is clear that relational problems -- the ones involving shape comparisons under different geometric transformations -- are the most difficult to solve for Deep Neural Networks. Thus, as in \cite{messina2019testing,messina2021solving}, we focus the attention on four of these problems: \textbf{Problem 1 (P.1)} - detecting the very same shapes, randomly placed in the image, having the same orientation and scale; \textbf{Problem 5 (P.5)} - detecting two pairs of identical shapes, randomly placed in the image. \textbf{Problem 20 (P.20)} - detecting the same shape, translated and flipped along a randomly chosen axis; \textbf{Problem 21 (P.21)} - detecting the same shape, randomly translated, orientated, and scaled. Positive and negative samples from each of these visual problems are shown in Figure \ref{fig:svrt-examples}.

\subsection{Setup}
For the upstream CNN processing the pixel-level information, we used a 4-layer \textit{Steerable CNN} \cite{e2cnn}. A Steerable CNN describes E(2)-equivariant (i.e., rotation- and reflection-equivariant) convolutions on the image plane $\mathbb{R}^2$;
in contrast to conventional CNNs, E(2)-equivariant models are guaranteed to generalize over such transformations other than simple translation and are therefore more data-efficient. In the ablation study in Section \ref{sec:ablation}, we will give more insights on the role of Steerable CNNs over standard CNNs in solving the same-different task.

We forged two different versions of the RViT, a \textit{small} and a \textit{large} version, having the same structure but a different number of hidden neurons in the core layers: the small RViT produces 256-dimensional keys, queries, and values and outputs 256-dimensional visual features from the CNN, while the large RViT has these two parameters set to 512. We used the Adam optimizer; after a minor hyper-parameter tuning, we set the learning rate for all the experiments to 1e-4, and the number of attention heads to 4; we let the models train for 200 epochs, decreasing the learning rate to 1e-5 after 170 epochs. We tested the models using the snapshot with the best accuracy measured on the validation set.

In order to better compare with the ResNet-50 experiments in \cite{borowski2019notorious}, we also tried to use as up-stream CNN the first two or three layers of a ResNet-50 pre-trained on ImageNet.
For the image resolution, we mainly used $N=16$, outputting $16\times16$ visual tokens from the CNN. During the pre-training experiments, instead, we used $N=8$ for accommodating the output feature map resolution of the pre-trained model and also for performance reasons. During training, we set the maximum time steps $T=9$. 

We collected results using both 28k training images, following  \cite{borowski2019notorious}, and 400k training images, for comparing our proposed architectures with convolutional networks trained in \cite{messina2019testing,messina2021solving}.
We used 18k images both for validation and testing. The images were generated with the SVRT original code, available online\footnote{\url{https://fleuret.org/git-tgz/svrt}}. 

\subsection{Results}

\newcolumntype{L}{>{\raggedright\arraybackslash}p{2.6cm}}
\newcolumntype{C}{>{\centering\arraybackslash}p{0.95cm}}
\begin{table}[t]
\begin{center}
\caption{Accuracy (\%) of our method, trained from-scratch, with respect to the baselines. \#pars indicate the number of free parameters of the model.}
\vspace{2mm}
\begin{tabular}{LCCCCCCCCC}
\toprule
& \multicolumn{4}{c}{400k training samples} & \multicolumn{4}{c}{28k training samples} & \\
\cmidrule(lr){2-5} \cmidrule(lr){6-9}
\textbf{Model} & \makecell{\textbf{P.1}\\$\uparrow$} & \makecell{\textbf{P.5}\\$\uparrow$} & \makecell{\textbf{P.20}\\$\uparrow$} & \makecell{\textbf{P.21}\\$\uparrow$} & \makecell{\textbf{P.1}\\$\uparrow$} & \makecell{\textbf{P.5}\\$\uparrow$} & \makecell{\textbf{P.20}\\$\uparrow$} & \makecell{\textbf{P.21}\\$\uparrow$} & \makecell{\#pars\\$\downarrow$}\\
\midrule
RN \cite{Santoro2017RelationNetworks} & \textcolor{gray}{50.0} & \textcolor{gray}{50.0} & \textcolor{gray}{50.0} & \textcolor{gray}{50.0} & \textcolor{gray}{50.0} & \textcolor{gray}{50.0} & \textcolor{gray}{50.0} & \textcolor{gray}{50.0} & 0.4M\\
ViT \cite{dosovitskiy2020image} & \textcolor{gray}{50.0} & \textcolor{gray}{50.0} & \textcolor{gray}{50.0} & \textcolor{gray}{50.0} & \textcolor{gray}{50.0} & \textcolor{gray}{50.0} & \textcolor{gray}{50.0} & \textcolor{gray}{50.0} & 26M \\
ResNet-18 \cite{messina2021solving} & 99.2 & \textbf{99.9} & 95.5 & 96.2 & 99.2 & 98.4 & 93.7 & \textcolor{gray}{50.0} & 11M\\
ResNet-50 \cite{borowski2019notorious} & - & - & - & - & 95.4 & 89.9 & 92.9 & 72.6 & 23M \\
DenseNet-121 \cite{messina2021solving} & 99.6 & 98.2 & 94.2 & 95.1 & 73.9 & 54.7 & 94.4 & \textbf{85.8} & 6.9M\\
CorNet-S \cite{messina2021solving} & 96.9 & 96.8 & 95.0 & \textbf{96.9} & 98.8 & 97.1 & 92.3 & 82.5 & 52M\\
\midrule
RViT-small & \textbf{99.9} & 99.4 & \textbf{98.9} & 95.7 & \textbf{99.6} & 98.0 & 93.9 & 78.6 & \textbf{0.9M} \\
RViT-large & \textbf{99.9} & 99.0 & 98.8 & 96.4 & \textbf{99.6} & \textbf{99.3} & \textbf{95.3} & 77.8 & 3.1M \\
\bottomrule
\end{tabular}
\label{tab:from-scratch}
\end{center}
\end{table}

\newcolumntype{L}{>{\raggedright\arraybackslash}p{3cm}}
\newcolumntype{C}{>{\centering\arraybackslash}p{1.7cm}}
\begin{table}[t]
\begin{center}
\caption{Accuracy (\%) of RViT-small, with the first layers of a ResNet-50 pre-trained on ImageNet, with respect to the full ResNet-50 baseline. In ResNet-50/11 we kept the first 11 layers, while in ResNet-50/23 the first 23.}
\vspace{2mm}
\begin{tabular}{LCCCCC}
\toprule
\textbf{Model} & \makecell{\textbf{P.1}\\$\uparrow$} & \makecell{\textbf{P.5}\\$\uparrow$} & \makecell{\textbf{P.20}\\$\uparrow$} & \makecell{\textbf{P.21}\\$\uparrow$} & \makecell{\#pars\\$\downarrow$}\\
\midrule
ResNet-50 \cite{borowski2019notorious} & 99.5 & 98.7 & 98.9 & \textbf{92.5} & 23M \\
RViT ResNet-50/11 & 99.6 & 98.6 & 94.5 & 91.6 & \textbf{2.3M} \\
RViT ResNet-50/23 & \textbf{99.7} & \textbf{99.7} & \textbf{99.4} & 85.2 & 9.5M \\
\bottomrule
\end{tabular}
\label{tab:pretrained}
\end{center}
\end{table}

We compared our model with other key architectures: the Relation Network (RN) \cite{Santoro2017RelationNetworks} which by design should be able to correlate distant zones of the image; the Vision Transformer (ViT) \cite{dosovitskiy2020image} which recently achieved remarkable performance on classification tasks, although it is very data-hungry, and some state-of-the-art convolutional models --- ResNet18, ResNet50, CorNet-S and DenseNet121 --- trained on the same task in \cite{borowski2019notorious,messina2019testing,messina2021solving}. Notably, CorNet-S also implements feedback connections, although it is much more complex, in terms of number of parameters, than our RViT architecture.

Looking at Table \ref{tab:from-scratch}, we can see how neither the Relation Network nor the ViT converges on the four visual problems, for both 400k and 28k data regimes. The ViT probably needs more architectural inductive biases to understand the rules, while the relational mechanism of Relation Network is probably too simple for understanding the objects in the image and their relationships. Instead, our RViT model can obtain very competitive results on all tasks and on both data regimes, often outperforming the baselines. Noticeably, the RViT-small can learn all the four problems using only 0.9M free parameters, about 8 times fewer parameters than the smallest convolutional network able to solve the task (DenseNet121). This suggests that the model has the correct structure for understanding the visual problems, without having the possibility to memorize the patterns. 

In Table \ref{tab:pretrained}, we instead report the accuracy of the small RViT model, where the upstream path is pre-trained on the classification task on ImageNet, following the work in \cite{borowski2019notorious}. Even in this case, the RViT achieves competitive results, but with much fewer free parameters and using only a slice -- the first 11 and 23 layers -- of the pre-trained ResNet-50 architecture.

\subsection{Ablation Study}
Following, we report some in-depth analysis of the RViTs performed with 28k training images.
\label{sec:ablation}

\subsubsection{The role of Recurrent Connections and Steerable Convolution}
In Table~\ref{tab:ablations}, we experimented with some variations of the RViT to understand the roles of recurrent connections and the employed 4-layers steerable CNN. The basic configuration is Conv. ViT, which is the same as the standard ViT from \cite{dosovitskiy2020image} but with an upstream CNN as the visual feature extractor. In contrast to the original ViT formulation, the Conv. ViT can improve significantly on P.1, P.20, and P.21, moving away from the chance accuracy. However, the most significant jump in accuracy happens when recurrent connections are introduced (Conv. RViT). In this case, the same model can learn all the visual problems, with an improvement of 67\% on P.1 and 7\% on P.21. 
Another improvement is obtained when using the Steerable CNNs \cite{e2cnn}. This kind of CNN produces features equivariant to rotations and reflections. For this reason, it has a wider impact on P.20 and P.21, where shapes are reflected and rotated, respectively.

\begin{table}[t]
\begin{center}
\caption{Ablation study on Convolutional ViT (Conv. ViT), on Convolutional Recurrent ViT (Conv. RViT), and Equivariant Convolutional Recurrent ViT (Eq. Conv. RViT). The last one is the model effectively employed in Tables \ref{tab:from-scratch} and \ref{tab:pretrained}. Accuracy (\%) is in this case measured on the validation set.}
\vspace{2mm}
\begin{tabular}{LCCCC}
\toprule
\textbf{Model} & \makecell{\textbf{P.1}\\$\uparrow$} & \makecell{\textbf{P.5}\\$\uparrow$} & \makecell{\textbf{P.20}\\$\uparrow$} & \makecell{\textbf{P.21}\\$\uparrow$}\\
\midrule
Conv. ViT & 59.5 & \textcolor{gray}{50.0} & 88.5 & 62.5 \\
Conv. RViT & 99.9 & 99.0 & 93.9 & 66.8 \\
Eq. Conv. RViT & 99.8 & 99.4 & 95.6 & 77.3\\
\bottomrule
\end{tabular}
\label{tab:ablations}
\end{center}
\end{table}

\begin{figure}[t]
  \begin{subfigure}[b]{0.5\textwidth}
      \includegraphics[width=1\linewidth,trim=0.2cm 0 1.5cm 1cm,clip]{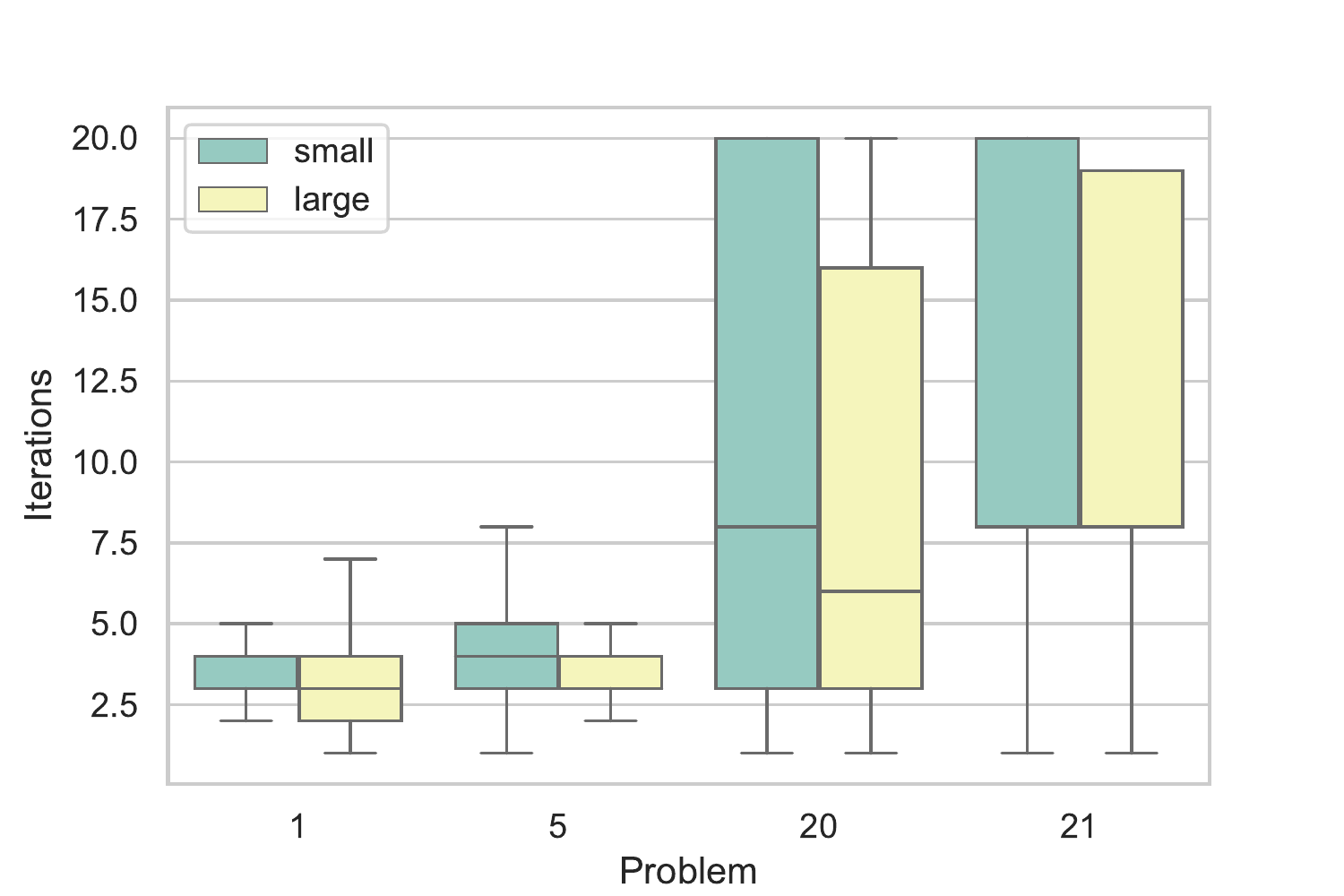}
      \caption{}
      \label{fig:iterations-smalllarge}
  \end{subfigure}
  \begin{subfigure}[b]{0.48\textwidth}
      \includegraphics[width=1\linewidth,trim=0.8cm 0 1.5cm 1cm,clip]{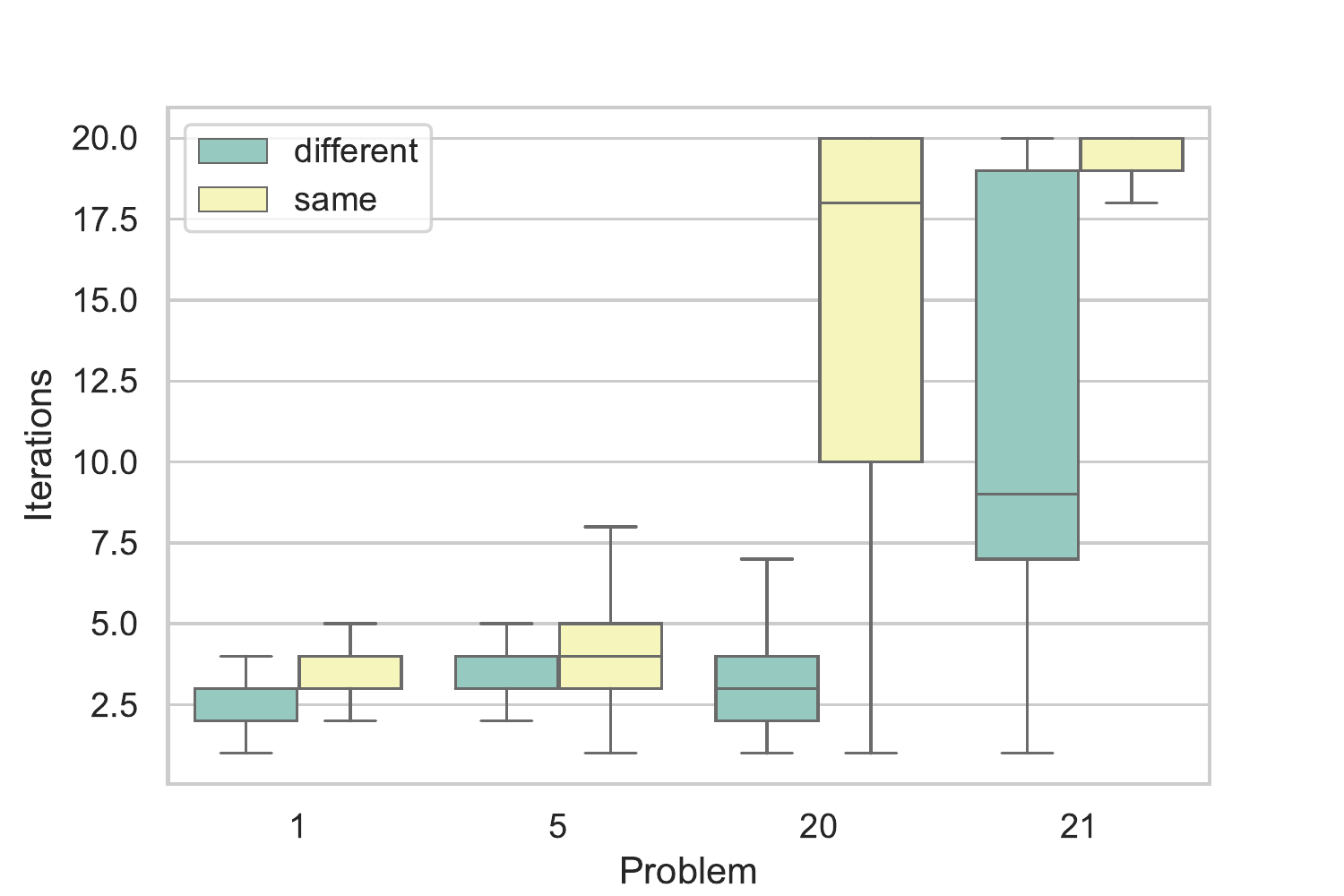}
      \caption{}
      \label{fig:iterations-samedifferent}
  \end{subfigure}
  \caption{The distribution of the best time step $\bar{t}$ grouped by (a) the two different RViT sizes (small, large), and (b) by the same-different label.}
  \label{fig:iterations}
\end{figure}

Recurrent connections seem to have critical importance. They highly regularize the model, making it more data-efficient and performing a dynamic iterative computation that procedurally refines both the previous internal representations and the previous predictions. To better appreciate this aspect, in Figure \ref{fig:iterations} we show the mean time step $\bar{t}$, for each problem, where the model reaches the maximum confidence. Interestingly, P.1 and P.5 reach the best confidence in few iterations, while the more challenging P.20 and P.21 need much more pondering before stabilizing. More in detail, it can be noticed that although there is not too much difference considering the size of the models (Figure \ref{fig:iterations-smalllarge}), the network seems majorly strained when the shapes are the \textit{same} (Figure \ref{fig:iterations-samedifferent}). This is reasonable: it is heavier to be sure that shapes coincide in every point, while it takes little to find even a single non-matching pattern to output the answer \textit{different}.

\subsubsection{Visualizing the Attention}
\begin{figure}[t]
  \centering
  \includegraphics[page=2,width=1\linewidth]{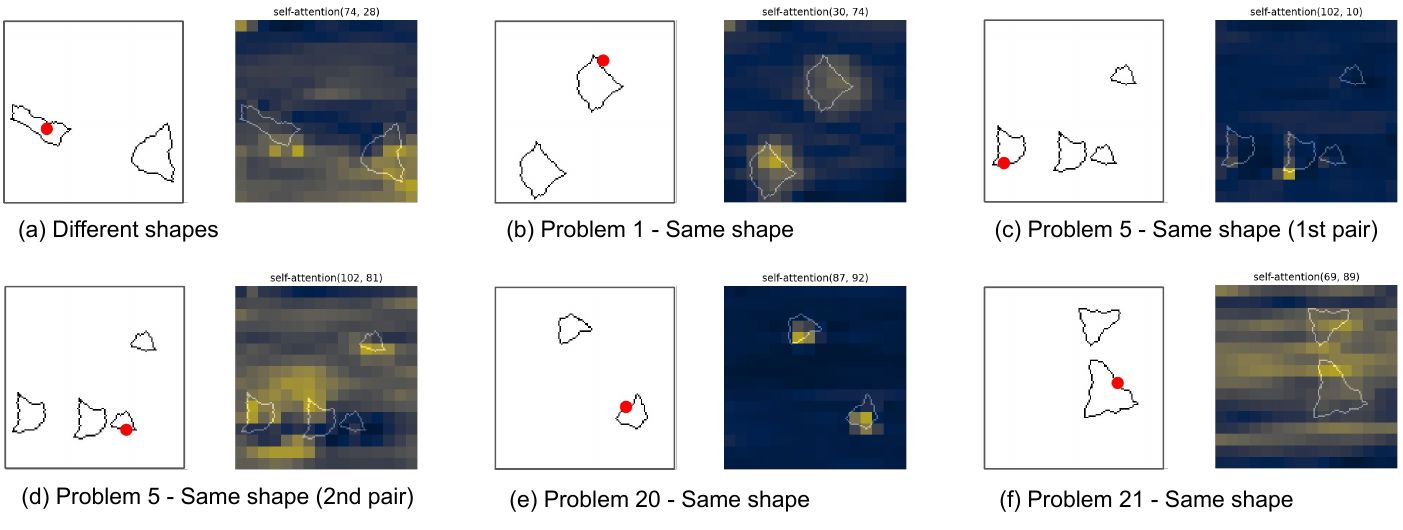}
  \caption{Attention visualization on the different visual problems. The red dot shows the point in space with respect to which the self-attention is computed.}
  \label{fig:attention} 
\end{figure}

\begin{figure}[t]
  \centering
  \includegraphics[page=3,width=0.9\linewidth]{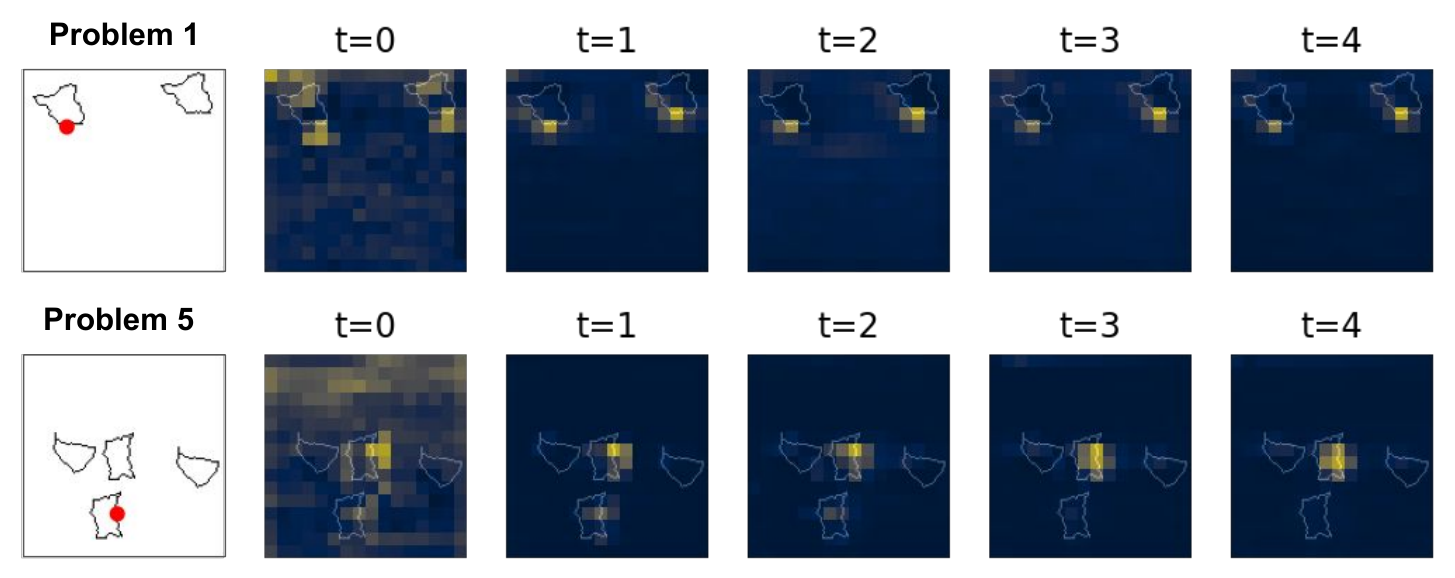}
  \caption{Evolving attention maps at different time steps.}
  \label{fig:attention-timesteps} 
\end{figure}

In Figure \ref{fig:attention}, we reported a visualization of the self-attention maps learned by the trained models, computed in specific points (marked with red dots) in the image, and by averaging the four attention heads. The $16\times16$ grid allows us to appreciate fine details; in particular, we can see what parts of the shapes the model is attending to for producing the final answer. In most cases, the model correctly attends the other shape in search of the corresponding edges. In some instances, the attention map is not so neat (e.g., in (d) and (f)), emphasizing the intrinsic complexity of the tasks.
Furthermore, in Figure \ref{fig:attention-timesteps} we report the evolving attention maps at different time steps. The map is initially very noisy, but it is slowly refined as the number of  iterations increases to create a stable representation.

\section{Conclusions}
In this work, we leveraged the power of Vision Transformer and recurrent connections to create the Recurrent Vision Transformer Model (RViT) capable of solving some of the most challenging same-different tasks from the SVRT dataset.
The experiments confirm the hypothesis that recurrent connections provide helps for understanding these visual problems, and the Transformer-like spatial attention enabled us to visualize what parts of the image the model is attending during the inference. The model outperforms the basic ViT model on this task, as well as other relation-aware architectures such as Relation Networks.
In the near future, this research could be extended to more challenging abstract reasoning tasks, such as RAVEN matrices, and a dynamic stopping mechanism can be included for probabilistically halting the computation after a stable prediction is obtained.


%
%

\bibliographystyle{splncs04}
\bibliography{biblio}

\end{document}